\PassOptionsToPackage{hyperfootnotes=false}{hyperref}
\makeatletter
\AddToHook{begindocument/before}{%
  \g@addto@macro\@title{\thanks{Accepted at GroundLM: Grounding Language Models:
    Learning Faithfully and Efficiently, a workshop at EMNLP 2026.}}%
  \hypersetup{pdfsubject={Accepted at GroundLM, EMNLP 2026 workshop}}%
}
\makeatother
\documentclass[11pt]{article}
\usepackage[final]{acl}
\usepackage{times}
\usepackage{latexsym}
\usepackage[T1]{fontenc}
\usepackage[utf8]{inputenc}
\usepackage{microtype}
\usepackage{inconsolata}
\usepackage{graphicx}
\usepackage{booktabs}
\usepackage{amsmath}
\usepackage{amssymb}
\usepackage{xcolor}
\usepackage{enumitem}
\title{Do {LLM} Attribution Metrics Transfer?\\Auditing Retrieval-Augmented Generation Evaluation Across Datasets and Constructs}

\author{
  Tianyu Ding \\ \texttt{tianyd@amazon.com} \\ Amazon Web Services
  \And
  Aditya Nannapaneni \\ \texttt{anannap@amazon.com} \\ Amazon Web Services
  \And
  \begin{tabular}[t]{@{}c@{}}\bfseries Juan Pablo De la Cruz\\\bfseries Weinstein\end{tabular}
  \\ \texttt{jcruam@amazon.com} \\ Amazon Web Services
}

\begin{document}
\maketitle

\begin{abstract}
Practice often treats automatic metrics for attribution in LLM retrieval-augmented generation as
interchangeable. We audit eight automatic scorers --- lexical, embedding, and BERTScore baselines
alongside entailment/grounding-trained models (clean and FEVER NLI, the checker MiniCheck) --- across
three evaluation constructs (provenance/topicality, generated-answer attribution, and fact-check
entailment), asking whether any scorer \emph{transfers}: stays within the $95\%$ confidence interval of
the best audited scorer on \emph{every} dataset of a multi-dataset construct. In the construct with the
most multi-dataset \textbf{human}-labeled coverage --- generated-answer attribution (AttributionBench's
four source datasets, $n{=}1610$, with independent HAGRID, $n{=}2150$) --- none of the audited automatic scorers does: the
per-dataset metric rankings invert (Kendall $\tau{=}-0.64$, $p{=}0.031$ on AttributedQA vs.\ LFQA), and an off-the-shelf NLI scorer that is best
on short-claim AttributedQA (AUROC $0.90$) collapses to AUROC $0.53$ (chance) on long-form LFQA, where
BERTScore wins ($0.91$); the reversal persists under the tested truncation settings. This instability has a concrete
decision cost: a naive ``best-on-average'' rule for choosing an evaluator \emph{fails}
leave-one-dataset-out (mean held-out regret $0.172$ AUROC, worse than fixing one scorer), so metric choice
should be validated on the target dataset rather than assumed from performance elsewhere. A prompt-based LLM judge avoids the
chance-level collapses the automatic scorers suffer (no LFQA collapse) but is not uniformly best, ${\sim}100\times$ costlier, and non-deterministic ---
relocating, not removing, the validation burden.
\end{abstract}

\section{Introduction}
Retrieval-augmented language models are increasingly evaluated for
\emph{grounding}: whether each claim in an answer is supported by the provided
evidence. A growing toolbox of automatic scores --- lexical overlap, sentence-embedding similarity,
NLI/entailment --- stands in for human grounding judgments, and practitioners often report them
interchangeably, as if ``the attribution metric went up'' meant the same thing regardless of
metric or benchmark. By ``LLM attribution metrics'' we mean metrics evaluating attribution in
LLM/RAG outputs (lexical, embedding, NLI/checker, and prompted LLM-judge variants); our core
audit is the cheaper \emph{automatic} scorers, with the prompted LLM judge as a boundary case.
We show this assumption is unsafe at two levels. Coarsely,
metric-family performance varies across the audited \emph{constructs} and datasets. More
sharply, \emph{within} a single construct the best metric is not stable across its
datasets, so no audited automatic scorer passes our core generated-answer attribution portability screen.
This is a claim about \emph{evaluator selection} at the item level (which scorer to trust on a given
dataset), not about the ranking of complete RAG systems, which we do not study (see \S\ref{sec:limitations}).

Concretely, we audit eight metrics --- off-the-shelf scores with no
attribution-specific fine-tuning (lexical Jaccard,
MiniLM and MPNet cosine, their fixed blend, BERTScore) and trained
entailment/grounding models (clean-MNLI NLI, FEVER NLI, and the purpose-built checker
MiniCheck \citep{tang2024minicheck}; a lexical/semantic/combined/NLI subset on the
provenance ranking task) --- across three evaluation constructs the literature
routinely conflates:
\textbf{(i) provenance/topicality} --- does a score rank a relevant passage top (ASQA
\citep{stelmakh2022asqa}, MS MARCO \citep{nguyen2016msmarco}, HotpotQA
\citep{yang2018hotpotqa}; passage-ranking labels, not sentence/claim-level support);
\textbf{(ii) generated-answer attribution} --- does a score predict \emph{human}
judgments of whether a generated answer sentence is supported by its cited evidence
(AttributionBench \citep{li2024attributionbench});
\textbf{(iii) fact-check entailment} --- does a score predict \emph{human}
support labels on short edited claims (VitaminC \citep{schuster2021vitaminc}).
At the coarse, cross-construct level metric-family performance differs on the audited benchmarks
(Table~\ref{tab:construct_audit}): the best off-the-shelf relevance score per provenance gate
reaches $0.73$--$0.97$ top-1 accuracy (within overlapping bootstrap CIs of NLI on MS~MARCO/HotpotQA,
though NLI is unreliable on ASQA at $0.585$), whereas
entailment-trained models score above all no-fine-tuning baselines on
fact-check claims (VitaminC, $0.81$--$0.89$ vs.\ surface ${\approx}0.60$ AUROC); this coarse contrast is
bounded to the audited benchmarks and partly confounded with domain shift (\S\ref{sec:limitations}). The sharper finding is
\emph{within} generated-answer attribution: across AttributionBench's four constituent
datasets the best metric is not stable (the per-dataset rankings invert, Kendall $\tau{=}-0.64$, $p{=}0.031$ on AttributedQA vs.\ LFQA).
A naive NLI scorer is best on AttributedQA (AUROC $0.90$) yet drops to $0.53$ (chance
$=0.50$) on LFQA, where BERTScore reaches $0.91$ --- though on a \emph{different} long-form
set (HAGRID) the same NLI scorer is best ($0.80$), so the failure is dataset-specific and
not explained by long-form \emph{answers} alone (long \emph{evidence} remains a candidate; \S\ref{sec:benchmark2}).

This matters because practitioners silently read success on one construct, or on one benchmark,
as evidence of grounding in general. Prior benchmarking work establishes that automatic
attribution metrics disagree with humans and that no single metric wins everywhere
\citep{honovich2022true, dziri2022begin, yue2023automatic, li2024attributionbench};
beyond that qualitative ``they disagree,'' our contribution is decision-relevant: an
operational \emph{portability} criterion, a leave-one-dataset-out \emph{regret} that prices
committing to one default, and a naive cross-dataset selector that \emph{fails} ($0.172$
regret) --- a validation discipline, on a controlled audit with human labels on the
in-domain attribution and fact-check sides and a clean separation from proxy-labeled
provenance (mechanism discovery is out of scope). This is an evaluation-resource/cautionary paper, not a new-metric paper.
Our central contribution is the cross-dataset evaluator-selection audit; the rest are supporting
diagnostics and boundary probes:

\begin{itemize}[leftmargin=*]
\item A \textbf{cross-dataset audit} (Section~\ref{sec:benchmark2},
  Table~\ref{tab:construct_audit}) showing metric-family differences across the audited constructs
  and datasets, but that \emph{within} in-domain answer attribution no single \emph{automatic}
  scorer is consistently near-best across datasets (rankings discordant, Kendall $W{=}0.07$, $p{=}0.029$): a raw MNLI scorer is best on
  AttributedQA ($0.90$) yet AUROC $0.53$ (chance $=0.50$) on the LFQA dataset --- though best again
  on the independent long-form HAGRID set ($0.80$) --- while a purpose-built checker
  (MiniCheck) has the lowest mean leave-one-dataset-out regret among the eight automatic scorers
  \emph{tested}, yet still leaves a measurable
  per-dataset regret (mean $0.044$ AUROC, CI $[0.014,0.072]$).
\item As supporting diagnostics, a training-free \textbf{sentence-unit provenance/topicality
  diagnostic} and a \textbf{confusable-evidence stress-test protocol}
  (Sections~\ref{sec:score}, \ref{sec:confusable}): under confusable evidence
  (high-overlap distractors from other examples), the semantic component separates
  co-provenant passages from distractors where lexical overlap is fooled. We benchmark
  an off-the-shelf NLI entailment head against it and find NLI benchmark-dependent,
  not uniformly best --- a construct-(i) diagnostic, not a general-purpose grounding metric.
\item On \textbf{validation discipline} (Section~\ref{sec:benchmark2}): a naive
  ``best-on-average'' selection rule fails leave-one-dataset-out (mean regret $0.172$,
  worse than the lowest-mean-regret fixed metric among those tested), so metric choice must be validated on the target dataset;
  and a \textbf{cross-family LLM-annotation} probe (Opus~4.8 + GPT-5.4, $n{=}160$) showing
  high inter-model but only moderate, dataset-dependent human agreement --- usable as a
  corroborative proxy only in the easiest (short-factoid) dataset, bounding off-the-shelf LLM label substitution.
\end{itemize}
An auxiliary rewrite-sensitivity probe appears in App.~\ref{app:probe}; it does not establish improved grounding.

In short, ``attribution metric'' names a family of signals capturing different constructs
(provenance, topical relatedness, human-judged support); metric suitability varies across the
audited settings, and \emph{within} answer attribution no audited automatic scorer passes our
portability screen across the tested datasets. Evaluations should report the dataset and construct they
target and prefer a checker validated on that dataset, not treat any single metric as a
universal proxy for grounding.

\section{Related Work}
\paragraph{Correctness vs.\ faithfulness.}
An answer can be correct while its citations do not support its claims
\citep{rashkin2023measuring, liu2023verifiability}: a faithfulness gap distinct from factual
correctness \citep{ji2023hallucination}, acute for long-form answers exact match cannot score
\citep{fan2019eli5} and worsened by uneven use of mid-context evidence \citep{liu2024lost}.

\paragraph{Attribution and citation quality.}
AIS \citep{rashkin2023measuring} formalizes source support, and audits find fluent
answers often carry unsupported citations \citep{liu2023verifiability}. ALCE
\citep{gao2023alce} introduced NLI-based citation recall/precision (with QAMPARI
\citep{amouyal2023qampari}); FActScore \citep{min2023factscore} decomposes into atomic
facts; Attribute-First \citep{slobodkin2024attributefirst} and LongCite
\citep{zhang2024longcite} produce attributable text; AttrScore \citep{yue2023automatic}
and RAGAS \citep{es2024ragas} judge support with LLM/entailment models. AttributionBench
\citep{li2024attributionbench} benchmarks such evaluators against human labels --- but,
crucially, only \emph{trained/LLM} evaluators, not the off-the-shelf lexical/embedding baselines
we find within overlapping CIs in-domain. These methods need entailment/LLM judges or modify
generation; our scorers include a deterministic sentence-unit one
\citep{reimers2019sbert}, and our contribution is the cross-dataset \emph{audit}.

\paragraph{RL for grounding.}
Prior work trains models to cite evidence via rewards, from verified-quote RLHF
\citep{menick2022gophercite} to citation rewards \citep{huang2024training}, gated
sufficiency/abstention rewards \citep{grace2026}, and verifiable-reward training
\citep{pala2025lessons}. These \emph{fix the model}; we \emph{diagnose the evaluation}, and our null result
(App.~\ref{app:probe}) is consistent with their report that answer-level metrics are hard to
move.

\paragraph{Distractors and robustness.}
Contextual distractors cause large drops \citep{lee2026noisybench}, especially
\emph{highly semantically related} ones \citep{wu2024irrelevant}, and grounding can fail
even in instruction-tuned models under adversarial conditions \citep{koneru2026pressure}. Our
confusable-evidence test (Section~\ref{sec:confusable}) is a controlled,
passage-ranking instance.

\section{A Sentence-Unit Provenance-Ranking Score}
\label{sec:score}
We score how well each sentence of an answer is \emph{matched} to a set of candidate
passages, for the provenance-ranking diagnostic of construct (i) (distinct from the
human-support task of constructs ii, iii). Let an answer be split into sentence units
$u_1,\dots,u_m$ and let the evidence be passages $p_1,\dots,p_k$. For a unit $u$ and
passage $p$ we compute a combined relevance score
\begin{equation}
\label{eq:score}
\begin{split}
s(u,p) = {}& 0.40\,\mathrm{lex}(u,p) + 0.45\,\mathrm{sem}(u,p) \\
           & {}+ 0.10\,h(u) + 0.05\,c(u,p),
\end{split}
\end{equation}
where $\mathrm{lex}$ is token Jaccard, $\mathrm{sem}$ is sentence-transformer cosine
\citep{reimers2019sbert} (\texttt{all-MiniLM-L6-v2}), the \emph{answer-hint} $h(u)$ is
the fraction of gold short answers in the unit, and the \emph{citation bonus} $c(u,p){=}1$
when $u$ cites $p$. A unit's score is $\max_p s(u,p)$. It is
\textbf{training-free} (only a small frozen encoder).

The score is a fixed \emph{provenance} diagnostic for construct (i), \textbf{not} a proposed
general-purpose grounding metric (far from best on the human-support tasks of
Section~\ref{sec:benchmark2}) and \textbf{not} the paper's central contribution: the
human-labeled cross-dataset audit of Section~\ref{sec:benchmark2}. We report the single
canonical weighting, not a tuned one. The answer-hint $h(u)$ is partly label-aware, but it does
not affect the confusable ranking: $h(u)$ is constant within a unit (so it cannot reorder that
unit's candidate passages), and the citation term $c(u,p){=}0$ throughout because ASQA gold
answers carry no inline citations. The confusable test thus ranks on lexical/semantic only and
the central result needs no gold-label access (ablated below).

\subsection{Provenance/topicality (construct i)}
\label{sec:gates}
The first construct asks whether a score routes an answer sentence to the passage it
came from. On ASQA we form a gold-reference vs.\ unadapted-model contrast: the gold
long-form answer vs.\ an unadapted \texttt{Qwen2.5-3B-Instruct} generation over the example's $\geq2$ passages
($623$ dev examples; the confusable test uses a stricter $\geq3$-passage filter, $339$).\footnote{Two
small ASQA checks (App.~\ref{app:gates}) are \emph{internal consistency} only (circular labels), so we
do not rely on them.} Holding
short-answer correctness roughly equal ($98$ examples within $0.15$ recall), the combined
provenance gap is $+0.160$, CI $[+0.133,+0.189]$ (full-dev $+0.164$), and is not a hint artifact
(zeroing the hint weight moves the matched gap only to $+0.153$, CI excluding $0$;
App.~\ref{app:component}). So provenance-sensitive scores move even when correctness does not
--- but this is provenance sensitivity, not human-judged support.

\paragraph{Confusable-evidence stress test.}
\label{sec:confusable}
To test the ``just lexical overlap'' objection, for each gold answer sentence we build a pool from the unit's
own ASQA passages plus $k{=}5$ \emph{confusable distractors} (highest lexical-overlap passages from
\emph{other} examples). The label is a \textbf{provenance proxy} (did the top passage come from the unit's
own example), not a human groundedness judgment. We report top-1 provenance accuracy and the \emph{fooled
rate} (a confusable distractor outranks every co-provenant passage); ranking uses lexical/semantic only.

\begin{table}[t]
\centering\small
\begin{tabular}{lcc}
\toprule
Scorer & Top-1 prov.\ acc. & Fooled rate \\
\midrule
Lexical-only & 0.775 & 0.225 \\
\textbf{Semantic-only} & \textbf{0.934} & \textbf{0.066} \\
Combined (canon.) & 0.919 & 0.081 \\
NLI entailment$^{\dagger}$ & 0.585 & 0.415 \\
\bottomrule
\end{tabular}
\caption{Confusable-evidence stress test on ASQA (top-1 averaged across four seeds; best in bold; semantic encoder \texttt{all-mpnet-base-v2}, $k{=}5$ distractors). Lexical / semantic / combined are on the full usable dev pool (339 ex / 1309 units). The combined$-$lexical gap is $+0.144$, 95\% CI $[+0.122,+0.165]$: all four independent-seed gaps ($+0.144,+0.144,+0.144,+0.143$) have CIs above $0$. $^{\dagger}$The NLI row is computed on the 200-example-per-seed subsample (3 seeds, $\approx$760 units each) for cost, so its absolute value is not directly pooled with the others; the lexical and semantic values on that same subsample are 0.786 and 0.938, so NLI's 0.585 is well below both on matched data. Ground truth is a passage-provenance proxy; distractors are highest-lexical-overlap passages from other examples. NLI = an off-the-shelf DeBERTa-v3 MNLI/FEVER/ANLI entailment head (passage$\rightarrow$sentence) on this ranking task, not full ALCE citation evaluation.}
\label{tab:confusable}
\end{table}

\paragraph{The semantic component survives; lexical is fooled; NLI does not help.}
On the full dev pool ($339$ examples, $1309$ units; semantic encoder \texttt{all-mpnet-base-v2}
for this confusable test), top-1 provenance accuracy is
$0.775$ lexical-only, $0.934$ semantic-only, $0.919$ combined (combined$-$lexical $+0.144$,
CI $[+0.122,+0.165]$, all four fresh-example seed gaps $>0$: $0.144, 0.144, 0.144, 0.143$); the fooled rate drops from
$22.5\%$ (lexical) to $8.1\%$ (combined) and $6.6\%$ (semantic). An off-the-shelf DeBERTa
MNLI/FEVER/ANLI scorer reaches only $0.585$ here --- below lexical, far below semantic --- a
construct mismatch (ASQA sentences \emph{synthesize} across passages; cf.\
\citealp{laban2022summac}), not a verdict on entailment in general. Sweeping the lexical weight $\alpha$
(Fig.~\ref{fig:ablation}, App.~\ref{app:component}) confirms the canonical blend
($\alpha\approx0.47$) is past the knee, so we do \textbf{not} claim it optimal; this is
specific to construct (i), where semantic carries the signal --- on human-judged support
(ii, iii) it is \emph{not} dominant.

\section{The Cross-Dataset Audit}
\label{sec:benchmark2}
We make our central point in two steps: coarsely, metric-family performance differs across
the audited constructs and datasets; more sharply, \emph{within} a construct the best metric stays
unstable across datasets. For this audit we use \emph{portable} (its ranking
\emph{transfers}) as one near-best screen, operational \emph{for metric selection}: a metric stays within the
$95\%$ CI of the best \emph{audited} metric on \emph{every} dataset of a construct. We do not
offer this as a universal definition of transfer, and our consequence is at the level of choosing an
evaluator, not ranking generators (system-level transfer is future work; see Limitations). In our data the same non-portability
reading is corroborated by discordant per-dataset metric rankings (concordance $W{=}0.07$, $p{=}0.029$; the AttributedQA-vs-LFQA ranking inverts, Kendall $\tau{=}-0.64$, $p{=}0.031$), a non-zero
leave-one-dataset-out regret (mean $0.044$ AUROC for the best fixed choice), paired-bootstrap
sign flips excluding zero (NLI$-$BERTScore $=-0.378$ on LFQA vs.\ $+0.213$ on AttributedQA),
the $\epsilon$-regret sweep (no audited metric transfers at $\epsilon\le0.05$; only MiniCheck at $\epsilon{=}0.10$), and the further sensitivity analyses in
App.~\ref{app:replication}. Dataset-dependent winners also occur in an independent replication
excluding the overlapping ExpertQA and LFQA datasets ($n{=}7336$, App.~\ref{app:replication}). We audit eight
metrics --- off-the-shelf scores with no attribution-specific fine-tuning (lexical Jaccard, MiniLM and MPNet cosine, their
blend, BERTScore \citep{zhang2020bertscore}) and trained models (clean non-FEVER MNLI NLI,
FEVER NLI, and the purpose-built checker MiniCheck \citep{tang2024minicheck}) --- on the
two support constructs, and a lexical/semantic/combined/NLI subset on the provenance
ranking task; we never pool the three constructs:
\textbf{(i) provenance/topicality} (ASQA, MS MARCO, HotpotQA: rank candidate passages and
check the top one; ASQA uses a provenance proxy, MS MARCO and HotpotQA use human
passage-level labels, none sentence/claim-level support);
\textbf{(ii) generated-answer attribution} with \emph{human} support labels
(AttributionBench \citep{li2024attributionbench}: is a generated answer sentence
attributable to its cited evidence?);
\textbf{(iii) fact-check entailment} with \emph{human} support labels
(VitaminC \citep{schuster2021vitaminc}: is a short claim supported by its evidence?).
For (ii, iii) we score each (claim/sentence, evidence) pair and report AUROC (clustered
bootstrap CIs); the clean NLI baseline uses a non-FEVER MNLI model to avoid VitaminC
leakage, while the provenance ranker (c) uses an off-the-shelf DeBERTa MNLI/FEVER/ANLI head
(so the NLI \emph{column} differs by construct). Table~\ref{tab:construct_audit} is the headline.

\begin{table*}[t]
\centering\footnotesize
\setlength{\tabcolsep}{4pt}
\resizebox{\textwidth}{!}{%
\begin{tabular}{@{}llcccccccc@{}}
\toprule
& & \multicolumn{5}{c}{\emph{off-the-shelf (no fine-tuning)}} & \multicolumn{3}{c}{\emph{trained}} \\
\cmidrule(lr){3-7}\cmidrule(lr){8-10}
Construct & Benchmark (label type) & Lex & MiniLM & MPNet & Comb & BERTSc & NLI$_{\mathrm{cl}}$ & NLI$_{\mathrm{fe}}$ & MiniChk \\
\midrule
\multicolumn{10}{@{}l}{\emph{(a) Generated-answer attribution} --- \textbf{human} support labels, AUROC} \\
\;\; AttributedQA (short claims) & AttributionBench \citep{li2024attributionbench} & 0.634 & 0.813 & 0.766 & 0.807 & 0.692 & \textbf{0.904} & 0.897 & 0.884 \\
\;\; LFQA (long-form)            & AttributionBench & 0.802 & 0.739 & 0.771 & 0.764 & \textbf{0.909} & 0.531 & 0.720 & 0.852 \\
\;\; Stanford-GenSearch         & AttributionBench & 0.818 & 0.800 & 0.779 & 0.828 & \textbf{0.838} & 0.822 & 0.765 & 0.783 \\
\;\; ExpertQA (underspecified)   & AttributionBench & 0.539 & 0.608 & \textbf{0.618} & 0.603 & 0.569 & 0.565 & 0.608 & 0.577 \\
\;\; HAGRID (long-form, indep.)  & \citet{kamalloo2023hagrid} & 0.752 & 0.719 & 0.750 & 0.722 & 0.793 & \textbf{0.800} & 0.787 & 0.791 \\
\midrule
\multicolumn{10}{@{}l}{\emph{(b) Fact-check entailment} --- \textbf{human} support labels, AUROC} \\
\;\; VitaminC (short claims) & VitaminC \citep{schuster2021vitaminc} & 0.619 & 0.613 & 0.609 & 0.628 & 0.596 & 0.811 & \textbf{0.894} & 0.810 \\
\midrule
\multicolumn{10}{@{}l}{\emph{(c) Provenance / topicality} --- passage-level labels, top-1 acc.} \\
\;\; provenance proxy        & ASQA \citep{stelmakh2022asqa}     & 0.775 & --- & \textbf{0.934} & 0.919 & --- & 0.585 & --- & --- \\
\;\; human \texttt{is\_selected} & MS MARCO \citep{nguyen2016msmarco} & 0.683 & 0.646 & --- & \textbf{0.726} & --- & 0.708 & --- & --- \\
\;\; human support-facts     & HotpotQA \citep{yang2018hotpotqa} & 0.959 & 0.938 & --- & 0.972 & --- & \textbf{0.979} & --- & --- \\
\bottomrule
\end{tabular}}
\caption{\textbf{Benchmark-by-benchmark audit; constructs are reported separately and never pooled.}
The main multi-dataset non-portability result is the four-source AttributionBench audit in (a);
HAGRID, VitaminC, and the provenance benchmarks are independent boundary/contrast checks, not
evidence for a universal metric ranking. AUROC for the
support tasks (a, b; binary support against \emph{human} sentence/claim labels) and
top-1 \emph{passage-ranking} accuracy (not AUROC) for provenance (c; passage-level labels --- a provenance proxy for
ASQA, human passage-relevance for MS MARCO/HotpotQA --- which are \emph{not}
sentence/claim-level support judgments); best per row in bold. For the human-support tasks (a, b) NLI$_{\mathrm{cl}}$ is the
\emph{clean, non-FEVER} MNLI model (\texttt{roberta-large-mnli}) and
NLI$_{\mathrm{fe}}$ is FEVER-trained; MiniChk is MiniCheck
\citep{tang2024minicheck}. For the human-support tasks (a, b), Comb is the two-term blend
$0.47\cdot\text{lex}+0.53\cdot\text{sem}$ over each (claim/sentence, evidence) pair (the
answer-hint and citation terms of Eq.~\eqref{eq:score} do not apply --- support pairs carry
no gold answer or inline citation --- and are omitted); MiniLM is the semantic encoder there.
The provenance NLI column (c) is the off-the-shelf
DeBERTa-v3 MNLI/FEVER/ANLI head used as a ranker (Section~\ref{sec:confusable}). \textbf{Coarsely}, metric-family performance differs across the audited settings:
entailment-trained models score above all no-fine-tuning baselines on fact-check (b),
while off-the-shelf relevance is within overlapping CIs of NLI on provenance (c). \textbf{But within one construct} (a), the best metric is not stable
across the four AttributionBench source datasets (the per-dataset metric rankings are discordant: Kendall $W{=}0.07$, $p{=}0.029$; the AttributedQA-vs-LFQA ranking inverts, $\tau{=}-0.64$;
HAGRID is an independent long-form set): the clean MNLI scorer is best on short-claim AttributedQA
($0.904$) yet AUROC $0.531$ (chance $=0.50$) on long-form LFQA, where BERTScore wins
($0.909$). The LFQA collapse persists at both tested NLI token limits ($256$, $512$).
Length-controlled analyses leave residual dataset differences but do not isolate their cause.
MiniCheck has the lowest mean leave-one-dataset-out regret
\emph{among those tested} (AttributionBench mean AUROC $0.774$; regret $0.044$) but still not uniformly
best. ExpertQA is an underspecified dataset (oracle best-of-eight only $0.618$). The
semantic provenance ranker (c) is \texttt{all-mpnet-base-v2} for the ASQA confusable test
(Section~\ref{sec:confusable}; hence its $0.934$ sits in the MPNet column) and
\texttt{all-MiniLM-L6-v2} for MS MARCO/HotpotQA; BERTScore, NLI$_{\mathrm{fe}}$, and
MiniCheck are AUROC support scorers and are not run as provenance \emph{rankers} (c).
Human-labeled \emph{support} (a, b) and proxy-labeled
\emph{provenance} (c) are distinct constructs and are not pooled.}
\label{tab:construct_audit}
\end{table*}

\paragraph{Metric-family performance differs across the audited constructs and datasets.}
For fact-check entailment we have only one audited boundary dataset, VitaminC ($n{=}4000$ human
labels), where the entailment-trained models score above all no-fine-tuning baselines: FEVER
NLI $0.894$, clean NLI $0.811$, MiniCheck $0.810$, versus the surface metrics clustered near
$0.60$--$0.63$ (BERTScore $0.596$, lexical $0.619$, combined $0.628$). We treat this as a
one-dataset contrast on short edited claims, \emph{not} a general law about fact-checking. For
provenance/topicality we report only passage-ranking results (ASQA provenance proxy; human
passage-level labels for MS MARCO/HotpotQA), reaching $0.73$--$0.97$ top-1 accuracy (within
overlapping bootstrap CIs of NLI, which is however unreliable on ASQA at $0.585$); these labels
are distinct from sentence/claim-level support and are not pooled with the support tasks. The
coarse separation thus holds only loosely and only on these audited benchmarks.

\paragraph{But the ``attribution'' label is not a reliability boundary.}
AttributionBench ($n{=}1610$ human labels) aggregates four source datasets the literature
files under one task --- is a generated answer sentence attributable to its cited evidence?
--- and a practitioner who reads ``attribution metric'' expects a single applicable choice.
We make no claim that the four are cognitively identical (they differ sharply in evidence
length and structure, which we use below to characterize \emph{where} metrics fail); our
point is the practitioner-facing one: this shared label does \emph{not} predict which metric
to trust. The best metric is not stable across the four, and the evidence that isolates this
\emph{crossover} is rank-based: the per-dataset metric rankings are discordant (Kendall's
$W{=}0.07$; rank-permutation $p{=}0.029$), the AttributedQA-vs-LFQA ranking literally inverts
(Kendall $\tau{=}-0.64$, $p{=}0.031$), and the best-vs-runner-up sign flips
(NLI$-$BERTScore $=-0.378$ on LFQA vs.\ $+0.213$ on AttributedQA, CIs exclude $0$). A pure
dataset-difficulty (main) effect cannot produce a rank reversal. A two-way decomposition of the
AUROC matrix places $35\%$ of the cross-cell variance in the interaction residual (a residual
permutation that holds each dataset's difficulty fixed is borderline, $p{=}0.053$), while most of the raw
cross-cell spread ($61\%$) is dataset-difficulty main effect: some datasets are simply harder for
every metric. (An exchangeability permutation that shuffles the dataset assignment of per-example
pairs rejects at $p<0.002$, but that test conflates the main effect with the interaction, so we
rely on the rank-based statistics above.) The clean MNLI scorer is the best
metric on short-claim AttributedQA ($0.904$; $+0.270$ over lexical Jaccard, paired
$95\%$ CI $[0.198,0.347]$) yet falls to AUROC $0.531$ (chance $=0.50$) on long-form LFQA,
where BERTScore reaches $0.909$ (MNLI$-$BERTScore $=-0.378$, CI $[-0.500,-0.238]$). This
collapse pins to one \emph{checkpoint}, not the entailment family: on the same LFQA,
FEVER-trained NLI scores $0.720$ and the checker MiniCheck $0.852$, so ``no single metric
transfers'' is about individual scorers, not a verdict on entailment models as a class. This
collapse persists when the NLI token limit increases from $256$ to $512$ (LFQA $0.531\!\to\!0.527$).
Length controls leave residual dataset differences: at matched evidence length ($89$--$180$ words) MNLI
still differs by source ($0.934$ AttributedQA vs.\ $0.597$ ExpertQA), and the within-source short-vs-long
drop ($+0.06$ LFQA, $+0.09$ ExpertQA) is far below the $0.38$ cross-dataset flip. A purpose-built checker (MiniCheck,
$2048$-token window) repairs the long-form failure (LFQA $0.852$) and has the lowest mean
leave-one-dataset-out regret \emph{among the metrics we tested} (AttributionBench mean
AUROC $0.774$), yet is not uniformly best, losing to BERTScore on Stanford-GenSearch
($-0.055$, CI $[-0.104,-0.010]$). Committing to one metric everywhere still incurs modest
per-dataset \emph{regret} vs.\ oracle selection (mean $0.044$, CI $[0.014,0.072]$) --- modest
in that it is well below the per-dataset sign-flips ($0.378$) that drive the headline, but its
CI excludes zero, so it is a real if small cost. ExpertQA is an underspecified dataset where every metric is weak (oracle best-of-eight only
$0.618$, barely above chance); three supplementary metrics agree (cross-encoder $0.58$,
MiniCheck-FT5 $0.60$, AlignScore $0.60$; App.~\ref{app:supp}), so the weakness is not isolated
to one scorer, and we lean no claim on it.

\paragraph{An external boundary: long-form alone does not predict the NLI failure.}
To test whether ``long-form'' explains the LFQA collapse, we score the full zoo on
\textbf{HAGRID} \citep{kamalloo2023hagrid}, an \emph{independent} long-form attribution
set (not an AttributionBench source) with human sentence/claim-level \texttt{attributable}
labels ($n{=}2150$). Here the clean MNLI scorer is instead the \emph{best} metric
($0.800$; BERTScore $0.793$, MiniCheck $0.791$, lexical $0.752$). HAGRID answers have
LFQA-comparable claim length but much shorter, explicitly \texttt{[n]}-cited evidence (short cited
quotes rather than LFQA's $\sim$320-word passages). This refutes the simplest reading: that NLI fails
whenever the \emph{answer} is long-form. It does \emph{not}, however, rule out long
\emph{evidence} (which HAGRID lacks) as the mediator; if anything, HAGRID's short cited
evidence is consistent with our evidence-length account below. Rankings are thus
dataset-specific even across superficially similar long-form settings; we cannot isolate the
causal latent characteristic from dataset identity here, and leave that to future work.

We also include BEGIN \citep{dziri2022begin} only as an appendix boundary check in a related
but different setting (knowledge-grounded \emph{dialogue} with response-level labels;
App.~\ref{app:begin}). Its pattern is consistent with dataset dependence, but it is not part
of the main multi-dataset audit, not used in the transfer criterion, and not co-equal with the
AttributionBench/HAGRID evidence. We caution
that the provenance results (construct i) use passage-level
provenance/relevance labels, not human sentence/claim-level support
judgments, and we do not pool them with
(ii, iii).

\paragraph{Naive cross-dataset metric selection does not generalize.}
Can one \emph{learn} a good default from observed datasets? Picking the metric with the best
mean AUROC on seen datasets fails leave-one-dataset-out over the four sources: mean realized
regret is $0.172$ AUROC, and the learned rule is on average $0.060$ AUROC \emph{worse} than
fixing the raw MNLI metric. The loss is
\emph{not} on the LFQA fold (there the rule picks MNLI and ties it); it comes from the other
folds --- most sharply AttributedQA-held-out, where the rule picks BERTScore ($0.69$) over
MNLI ($0.90$), a $-0.21$ swing --- so averaging over seen datasets actively mis-selects on
the held-out one. The actionable
consequence is a \emph{validation discipline}: specify the target construct; compare candidate
scorers on a representative sample with target-dataset human labels; then select and report
the scorer with its uncertainty and exclusions. Dataset properties alone do not establish a
selection rule here. These conclusions are bounded to the four datasets and eight metrics
audited here. A minimax rule minimizes the largest regret over the \emph{seen} datasets.
Using all four datasets, it selects MiniCheck, as does best-on-average (worst-case regret $0.058$).
Refitting on the three seen datasets in each leave-one-dataset-out fold gives the same choices
as best-on-average and the same $0.172$ mean held-out regret. This minimax rule does not
improve held-out performance; other robust selectors and ensembles are not evaluated.

\paragraph{When can LLM adjudication stand in for human labels?}
Since target-dataset human labels are often missing, we probe cross-family LLM adjudication:
two frontier models from \emph{different} families (Claude Opus~4.8, GPT-5.4) independently
annotate $160$ AttributionBench items ($40$/source), reported as \emph{LLM-adjudicated}, never
human/gold. The families agree with \emph{each other}
($\kappa{=}0.832$) but only moderately with \emph{human} labels ($\kappa{\approx}0.47$), and
that human-agreement is itself dataset-dependent (accuracy AttributedQA $0.93/0.88$ down to
ExpertQA $0.58/0.63$); on Stanford they agree strongly with each other ($\kappa{=}0.90$) yet
diverge from humans, so high inter-model agreement is \emph{not} evidence of validity. LLM
adjudication is thus a corroborative proxy only in the easiest (short-factoid) dataset.

\paragraph{Is a prompt-based LLM judge more stable across datasets?}
Our non-transfer finding is about \emph{automatic} scorers (including trained ones: clean/FEVER NLI,
MiniCheck). The natural rejoinder --- practitioners increasingly use a prompt-based \emph{LLM judge} --- we
test directly (Opus~4.8 as a scoring metric, Table~\ref{tab:llm_judge}): on the same four AttributionBench
sources it never drops to the chance-level collapses the strong automatic scorers suffer (AUROC $0.73$--$0.92$,
vs.\ clean MNLI $0.53$--$0.90$ and BERTScore $0.57$--$0.91$ on those four) and does \emph{not}
collapse on LFQA. The LLM judge \emph{avoids the collapses} here, but this is a boundary observation, not an escape
hatch: it is still not uniformly best, costs ${\sim}100\times$ more, and is non-deterministic; its AUROC is
computed over $408$ of $480$ sampled items with valid parsed scores ($15\%$ unscored).
Assigning $0.50$ to unscored items retains above-chance AUROC point estimates in all four datasets
(Table~\ref{tab:llm_judge}), but does not rule out selection bias. The judge relocates rather than
removes the validation burden.

\bigskip
We now detail the provenance benchmarks behind Table~\ref{tab:construct_audit}(c),
whose labels are passage-level (a provenance proxy for ASQA, \emph{human}
passage-relevance for MS MARCO and HotpotQA), \emph{not} sentence/claim-level support. On
\textbf{MS MARCO} v2.1 \citep{nguyen2016msmarco}
($11{,}836$ dev queries with $\geq3$ passages, a well-formed answer, and a
human-selected passage; ``combined'' is the gold-free $0.47\,\mathrm{lex}+0.53\,\mathrm{sem}$
blend), we rank each query's own passages and ask whether the top-1 is
\emph{human}-selected (\texttt{is\_selected}). Over $2000$ queries/seed $\times$3,
all scorers far exceed the random baseline ($0.104$): lexical $0.683$, semantic
$0.646$, NLI $0.708$, combined $0.726$ (combined and NLI within bootstrap noise, both
beating the components), so the score tracks \emph{human} passage relevance, a
construct-(i) sanity check, not grounded support. Under the confusable protocol
($n{=}300\times4$) lexical-only top-1 is $0.885$, semantic $0.995$, combined $0.996$
(combined$-$lexical $+0.111$, every per-seed CI excludes zero), and the added NLI
baseline reaches $0.952$ --- above lexical, below semantic. The same lexical-fooled,
semantic-survives ordering as ASQA, near-ceiling because MS MARCO is more extractive.

On \textbf{HotpotQA} \citep{yang2018hotpotqa} (distractor setting), with \emph{human}
supporting-fact annotations over ten candidate paragraphs ($1500$/seed $\times$3), all
scorers far exceed the random baseline $0.201$ (Table~\ref{tab:benchmark2}c); here, on
short multi-hop answers reusing the question's entities, \emph{lexical and NLI overtake
semantic}: the mirror image of ASQA.

\paragraph{Within provenance, no single component is universal; the blend is a
convenient baseline.}
The three \emph{provenance} benchmarks complicate any simple ``semantic wins'' summary: each single component has a dataset where it is weakest: semantic
similarity on short, entity-heavy multi-hop answers (HotpotQA), and lexical overlap
and off-the-shelf NLI on synthesized long-form answers (ASQA, where NLI falls to
$0.585$). The \emph{combined} score stays at or near the top across these
\emph{provenance} tasks (Tables~\ref{tab:construct_audit}c, \ref{tab:benchmark2};
on the human \texttt{is\_selected} gate it leads at $0.726$, and on the MS MARCO
confusable top-1 test it reaches $0.996$): within construct (i) only, the fixed blend is a
convenient provenance baseline that stays at or near the top across these retrieval datasets. This blend is \emph{not} a general-purpose \emph{grounding} metric: on the human-support tasks (constructs ii, iii) it
is far from best (Table~\ref{tab:construct_audit}), consistent with our headline
of non-portability in the core generated-answer attribution audit. The provenance behavior is
stable across encoders and distractor counts: \texttt{all-MiniLM-L6-v2} gives ASQA semantic accuracy
$0.919$--$0.939$ across $k{\in}\{3,8,10\}$ (vs.\ \texttt{all-mpnet-base-v2}'s $0.934$ at $k{=}5$), with the
combined$-$lexical gap positive ($+0.13$ to $+0.15$) throughout.

\section{Conclusion}
Under our operational near-best screen, no audited \emph{automatic}
scorer is consistently near-best across the datasets of generated-answer attribution, the
construct with the most multi-dataset human-labeled coverage (the four AttributionBench sources,
$n{=}1610$, with independent HAGRID, $n{=}2150$). Metric-family performance varies across the audited settings, but
within answer attribution the best metric is unstable across datasets (per-dataset rankings discordant,
Kendall $W{=}0.07$, $p{=}0.029$; the NLI scorer flips from best on AttributedQA to chance on LFQA),
and even the lowest-regret metric leaves measurable per-dataset regret.
A prompt-based LLM judge avoids the chance-level collapses (AUROC $0.73$--$0.92$, no LFQA collapse) but is costlier and
non-deterministic, relocating rather than removing the problem. Evaluations should report their dataset and
construct and validate the metric on them.

\section{Limitations}
\label{sec:limitations}
\textbf{Construct and domain shift are entangled.} Our largest-gap ``NLI wins'' contrast
($0.81$--$0.89$ vs.\ ${\approx}0.60$ AUROC) is also our only out-of-domain benchmark (VitaminC), so we cannot fully separate
construct from domain shift. The in-domain AttributionBench result (where the gap closes)
supports dependence on the evaluated setting, not an isolated construct effect or a claim
that entailment is intrinsically superior.
\textbf{Three sentence/claim-level human-support benchmarks, all English.} Only AttributionBench,
HAGRID, and VitaminC carry human \emph{sentence/claim-level} support labels (BEGIN,
App.~\ref{app:begin}, has human labels but at the \emph{response} level --- an external
boundary, not a main construct); all are English. Broader in-construct human labels (other
domains/languages) are the natural next step.
\textbf{Provenance uses passage-ranking labels, not the sentence-attribution task of (ii, iii).}
We pose construct (i) as passage ranking: ASQA uses a provenance proxy, while MS MARCO
(\texttt{is\_selected}) and HotpotQA (supporting-fact annotations) provide human relevance
labels. We use all three only to ask whether a score ranks a relevant passage top, \emph{not}
to predict sentence/claim-level support judgments. We keep these strictly separate from the human-support
constructs and never pool them.
\textbf{Correctness confound.} The full-dev provenance gap is partly correctness-confounded
(mitigated, not removed, by the matched subset).
\textbf{Item-level, not system-level.} Our consequence is at the level of choosing an evaluator under
dataset uncertainty (per-item AUROC and the failure of a naive cross-dataset selector); we do \emph{not}
show that this metric instability reverses \emph{system} rankings across generators. Whether item-level
non-transfer induces model-selection reversals is important future work.

\bibliography{references}

\appendix
\section{ASQA Internal Consistency Checks}
\label{app:gates}
For completeness we report two ASQA checks that we treat as internal consistency
checks, \emph{not} validation, because their labels are threshold heuristics derived
from the same lexical/semantic features under test (hence circular). \textbf{(1)
Gold-slice retrieval:} on a 20-example slice with 2--3 distractor passages, top-1
retrieval is $0.966$ combined / $0.915$ lexical / $0.966$ semantic against
heuristic ``gold'' labels. \textbf{(2) Matched-pair agreement:} on $10$ strong/decent
pairs the score's preference agrees $10/10$ with a lexical-overlap groundedness proxy
(gap $+0.185$). Both are corroborative only; the human-labeled constructs in
Table~\ref{tab:construct_audit} carry the paper's claims. The full gold-reference vs.\
unadapted-model gates and the answer-hint ablation are in Tables~\ref{tab:gates}--\ref{tab:hint}.
\begin{table}[t]
\centering\small
\setlength{\tabcolsep}{4pt}
\resizebox{\columnwidth}{!}{%
\begin{tabular}{@{}llll@{}}
\toprule
Gate & Metric & Result & Note \\
\midrule
1 & top-1 retrieval (comb.) & 0.966 & lex 0.915 / sem 0.966 \\
2 & matched-pair agreement & 10/10 & gap +0.185 \\
3 & full-dev gap ($n{=}623$) & +0.164 & CI[+0.152,+0.177] \\
3 & matched subset ($n{=}98$) & +0.160 & CI[+0.133,+0.189] \\
3 & evidence-prec.\ gap & +0.234 & --- \\
\bottomrule
\end{tabular}}
\caption{Three-gate internal-consistency check on ASQA dev. Labels are heuristic/proxy, not human. The correctness-matched subset (Gate 3, row 4) shows the largest gap.}
\label{tab:gates}
\end{table}
\begin{table}[t]
\centering\footnotesize
\setlength{\tabcolsep}{3.5pt}
\begin{tabular}{@{}lcc@{}}
\toprule
Gate-3 metric & Canonical & Hint $=0$ \\
& ($w_h{=}0.10$) & (ablated) \\
\midrule
Matched-subset gap & $+0.160$ & $+0.153$ \\
\quad 95\% CI & $[+0.133,+0.189]$ & $[+0.127,+0.180]$ \\
Full-dev gap & $+0.164$ & $+0.146$ \\
\quad 95\% CI & $[+0.152,+0.177]$ & $[+0.134,+0.157]$ \\
Evidence-prec.\ gap & $+0.234$ & $+0.205$ \\
\bottomrule
\end{tabular}
\caption{Answer-hint ablation on Gate 3 (ASQA full dev, $n{=}623$; matched subset
$n{=}98$). Removing the label-aware answer-hint term ($w_h{=}0$, other weights
fixed) leaves the grounding gap essentially intact, with the matched-subset CI
still excluding zero. The gap is not an artifact of the label-aware term.}
\label{tab:hint}
\end{table}

\section{Per-Component Ablation}
\label{app:component}
Table~\ref{tab:component} reports a leave-one-out ablation of the four score
components on the Gate-3 strong$-$decent gap. It complements the lexical-weight sweep
(Section~\ref{sec:confusable}) and the answer-hint ablation (Section~\ref{sec:gates}):
the lexical and semantic terms each carry a substantial share of the gap, while the
answer-hint and citation terms are minor, consistent with the main-text finding
that the score's signal does not depend on the label-aware hint.
\begin{table}[t]
\centering\small
\resizebox{\columnwidth}{!}{%
\begin{tabular}{lcc}
\toprule
Weighting & Matched gap & Full-dev gap \\
\midrule
Canonical (.40/.45/.10/.05) & $+0.160$ & $+0.164$ \\
\quad drop lexical & $+0.062$ & $+0.075$ \\
\quad drop semantic & $+0.097$ & $+0.101$ \\
\quad drop hint & $+0.153$ & $+0.146$ \\
\quad drop citation & $+0.168$ & $+0.170$ \\
Lexical-only & $+0.247$ & $+0.222$ \\
Semantic-only & $+0.139$ & $+0.138$ \\
\bottomrule
\end{tabular}}
\caption{Per-component leave-one-out on the Gate-3 strong$-$decent attribution gap
(ASQA full dev; matched subset $n{=}98$, full dev $n{=}623$; all matched CIs exclude
$0$). Dropping lexical or semantic shrinks the gap most (both contribute); dropping
the label-aware hint barely changes it ($+0.160\!\to\!+0.153$, confirming the gap is
not a hint artifact); dropping citation is harmless. The lexical-only and semantic-only
rows show each of those two components carries a gap on its own.}
\label{tab:component}
\end{table}

\begin{figure}[t]
\centering
\includegraphics[width=\columnwidth]{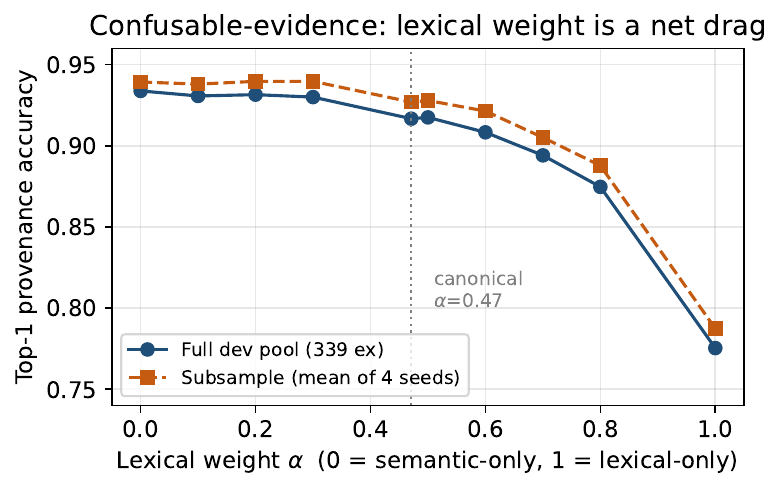}
\caption{Lexical-weight ablation under confusable evidence (construct i). Top-1
provenance accuracy is flat at low $\alpha$ and then declines as the lexical weight
$\alpha$ rises; the canonical blend ($\alpha\approx0.47$) is already past the knee.
Semantic-only ($\alpha{=}0$) is best.}
\label{fig:ablation}
\end{figure}

\section{A Grounding-Reward Stress Test: Do Answer-Level Metrics Detect a Large Rewrite?}
\label{app:probe}
To test whether answer-level metrics detect a large grounding-targeted rewrite, we ran a stress test. We
fine-tuned the generator (GRPO \citep{shao2024deepseekmath} on \texttt{Qwen2.5-\{3B,7B\}-Instruct})
with a reward that directly targets evidence grounding: it scores each answer sentence by
whether its claims are supported by the provided evidence and whether that support
\emph{counterfactually} depends on the evidence being present (rather than on the model's priors),
along with evidence coverage and epistemic commitment. Comparing the adapted model against the
\emph{unadapted} base (a base-vs-base$+$LoRA comparison, not SFT-vs-RL) on all $948$ ASQA dev
answers, at 3B the reward rewrites $89.9\%$ of generations, yet QA-EM, ROUGE-L, and citation count
stay flat. This is the paper's measurement point in miniature: a large, grounding-directed
intervention on the output leaves the answer-level metrics we test here flat, so those metrics are
insensitive to the kind of grounding-targeted rewrite this intervention produces. We use the reward
only as a probe, not as evidence that true grounding improved.

\paragraph{The metrics do not move.}
At 3B, QA-EM (corrected ASQA \texttt{str\_em}, any-alias match) is $0.299$ for the base and
$0.297$ for the adapted model (delta $-0.002$, $95\%$ paired-bootstrap CI $[-0.011,+0.008]$): no
statistically meaningful change. ROUGE-L moves $+0.006$ and citation count is flat. At 7B the
pattern repeats (QA-EM $0.377\!\to\!0.381$, CI $[-0.004,+0.012]$). Under a legacy substring metric
the 7B values are $0.240\!\to\!0.241$, the same flat pattern. We measure citation \emph{count},
which is flat; we did not regenerate NLI-based ALCE citation recall/precision \citep{gao2023alce}
(it requires a large entailment model and a different decoding setup), and we make no claim about it
here.

\paragraph{The intervention is not vacuous, and its scope is narrow.}
At 3B the reward alters $89.9\%$ of the $948$ generations (mean token Jaccard $0.59$; $81.2\%$
differ substantially, Jaccard $<0.9$), yet QA-EM, ROUGE-L, and citation count register essentially
nothing. We do \emph{not} claim the reward improved grounding (a single untuned run could be weak);
it shows only that these answer-level metrics are insensitive to a large grounding-targeted rewrite.
The scope is therefore narrow: the probe does not establish that these metrics would miss a
\emph{genuine} grounding improvement; it is motivation, not proof, and the audit does not depend
on it.

\paragraph{Setup.}
Table~\ref{tab:training} lists the LoRA and GRPO configuration, recovered from the saved adapter
configs and the training setup; training-time loss/reward trajectories were not retained.
Table~\ref{tab:ercr} reports the metric deltas. The regeneration uses greedy decoding: the flatness
reproduces, though absolute QA-EM differs from prior temperature-sampled prose.
\begin{table}[t]
\centering\small
\begin{tabular}{ll}
\toprule
Setting & Value \\
\midrule
\multicolumn{2}{l}{\emph{Base models}} \\
Backbone & \texttt{Qwen2.5-\{3B,7B\}-Instruct} \\
Adapter & LoRA (PEFT), \texttt{CAUSAL\_LM} \\
\midrule
\multicolumn{2}{l}{\emph{LoRA}} \\
Rank $r$ & $8$ \\
$\alpha$ & $16$ \\
Dropout & $0.0$ \\
Target modules & q,\,k,\,v,\,o,\,gate,\,up,\,down\,\texttt{\_proj} \\
\midrule
\multicolumn{2}{l}{\emph{GRPO (TRL GRPOTrainer)}} \\
Configuration & DAPO-style \\
Generations / prompt & $4$ \\
Max completion length & $512$ tokens \\
Input truncation & $2048$ tokens \\
Reward weights & fixed (not tuned) \\
\midrule
\multicolumn{2}{l}{\emph{Held-out evaluation}} \\
Decoding & greedy \\
Eval set & full ASQA dev ($948$) \\
\bottomrule
\end{tabular}
\caption{Grounding-reward training and evaluation configuration, recovered from the saved
LoRA adapter configs and the GRPO setup. We report these for methodological
transparency; the reward is a probe (App.~\ref{app:probe}), not a proposed method. Training-time
loss/reward trajectories were not retained with the saved adapters and are not
reported.}
\label{tab:training}
\end{table}

\begin{table}[t]
\centering\small
\begin{tabular}{llll}
\toprule
Model & QA-EM & ROUGE-L & Cites \\
\midrule
\multicolumn{4}{l}{\emph{3B}} \\
base (unadapted) & 0.299 & 0.219 & 1.05 \\
{}+ grounding reward & 0.297 & 0.225 & 1.04 \\
\multicolumn{4}{l}{$\Delta$QA-EM -0.002, CI[-0.011,+0.008]} \\
\midrule
\multicolumn{4}{l}{\emph{7B}} \\
base (unadapted) & 0.377 & 0.191 & 1.83 \\
{}+ grounding reward & 0.381 & 0.193 & 1.83 \\
\multicolumn{4}{l}{$\Delta$QA-EM +0.004, CI[-0.004,+0.012]} \\
\bottomrule
\end{tabular}
\caption{Grounding-reward stress test on full ASQA dev (948 ex, greedy decoding). QA-EM is corrected ASQA \texttt{str\_em}. The reward produces no statistically meaningful change at either scale.}
\label{tab:ercr}
\end{table}

\section{Supplementary Metric Families}
\label{app:supp}
Beyond the eight metrics of Table~\ref{tab:construct_audit}, we ran three additional
supplementary metrics on the AttributionBench sources to check that the per-source pattern
is not specific to our metric set: a cross-encoder reranker (\texttt{ms-marco-MiniLM-L6}),
MiniCheck-FT5 (\texttt{flan-t5-large}), and AlignScore-large \citep{zha2023alignscore}, a
metric purpose-built for factual consistency. Table~\ref{tab:supplementary} shows all
three reproduce the pattern, and in particular all are near chance on ExpertQA (AUROC
$0.58$--$0.60$, chance $=0.50$),
so the weak performance there is not isolated to the originally audited scorers.
\begin{table}[t]
\centering\footnotesize
\setlength{\tabcolsep}{3pt}
\resizebox{\columnwidth}{!}{%
\begin{tabular}{@{}lccc@{}}
\toprule
Dataset & Cross-enc. & MiniChk-FT5 & AlignScore \\
\midrule
AttributedQA        & 0.867 & 0.896 & \textbf{0.901} \\
LFQA                & 0.853 & \textbf{0.859} & 0.811 \\
Stanford-GenSearch  & \textbf{0.836} & 0.795 & 0.784 \\
ExpertQA            & 0.580 & 0.603 & 0.596 \\
\bottomrule
\end{tabular}}
\caption{\textbf{Three supplementary metrics reproduce the per-source pattern}
(AUROC on AttributionBench human attribution labels): a cross-encoder reranker
(\texttt{ms-marco-MiniLM-L6}), MiniCheck-FT5 (\texttt{flan-t5-large}), and AlignScore-large
\citep{zha2023alignscore}. All three reach AUROC $0.867$--$0.901$ on AttributedQA, weaker but usable on
LFQA/Stanford, and weak on ExpertQA (AUROC $0.58$--$0.60$) --- matching the eight metrics of
Table~\ref{tab:construct_audit}. ExpertQA is weak for every scorer we tried (AUROC $0.58$--$0.62$, only
just above chance), so its low ceiling is not isolated to any single metric.}
\label{tab:supplementary}
\end{table}

\begin{table}[t]
\centering\small
\begin{tabular}{lcc}
\toprule
\multicolumn{3}{l}{\emph{(a) Human passage-relevance gate} (top-1 \texttt{is\_selected})} \\
Scorer & Top-1 acc. & vs.\ random \\
\midrule
Random passage & 0.104 & --- \\
Lexical-only & 0.683 & $+0.579$ \\
Semantic-only & 0.646 & $+0.542$ \\
NLI entailment & 0.708 & $+0.604$ \\
\textbf{Combined (canon.)} & \textbf{0.726} & $+0.622$ \\
\midrule
\multicolumn{3}{l}{\emph{(b) MS MARCO confusable test} (top-1 provenance)} \\
Scorer & Top-1 acc. & Fooled \\
\midrule
Lexical-only & 0.885 & 0.115 \\
Semantic-only & 0.995 & 0.005 \\
\textbf{Combined (canon.)} & \textbf{0.996} & \textbf{0.004} \\
NLI entailment$^{\dagger}$ & 0.952 & 0.048 \\
\midrule
\multicolumn{3}{l}{\emph{(c) HotpotQA human supporting-fact gate}} \\
Scorer & Top-1 acc. & vs.\ random \\
\midrule
Random paragraph & 0.201 & --- \\
Lexical-only & 0.959 & $+0.758$ \\
Semantic-only & 0.938 & $+0.737$ \\
NLI entailment & 0.979 & $+0.778$ \\
\textbf{Combined (canon.)} & \textbf{0.972} & $+0.771$ \\
\bottomrule
\end{tabular}
\caption{Additional benchmarks (best in bold). \textbf{(a)} MS MARCO human
passage-relevance gate: rank a query's retrieved passages per answer sentence;
ground truth is the \emph{human} \texttt{is\_selected} label (passage-level, not
sentence/claim-level support), $2000$ queries $\times$ 3 seeds. \textbf{(b)} MS MARCO
confusable-evidence test (Section~\ref{sec:confusable} design), $n{=}300\times4$
seeds: the combined/semantic advantage over lexical holds (gap $+0.111$; four-seed
gaps $+0.097$ to $+0.130$, all CIs ${>}0$). \textbf{(c)} HotpotQA distractor-setting
human supporting-fact gate, $1500\times3$ seeds: rank a human-supporting paragraph top
(2-hop, so $\approx2$ supporting among 10; random $0.201$). On these extractive/multi-hop answers lexical and NLI
overtake semantic --- the mirror image of ASQA --- yet the combined score stays
near-best on every benchmark. $^{\dagger}$NLI is a separate $n{=}200\times3$ subsample.}
\label{tab:benchmark2}
\end{table}

\section{BEGIN: Appendix Boundary Check (Dialogue, Response-Level Labels)}
\label{app:begin}
As an external, different-domain check (knowledge-grounded \emph{dialogue} rather than
QA), we score the metric zoo on BEGIN \citep{dziri2022begin}, $n{=}11{,}466$ system
responses with human ``fully attributable'' vs ``not'' labels (we drop the ``generic''
class), using knowledge as evidence and the response as the claim. Table~\ref{tab:begin}
shows the grounding-trained family (clean MNLI, MiniCheck) has top AUROC on every dataset here ---
the opposite of LFQA, where clean MNLI collapsed. This is a boundary condition that
sharpens, rather than contradicts, non-portability: an otherwise top-scoring family still has a
dataset (LFQA) where it fails, so no single metric is the lowest-regret fixed choice across datasets. We
treat BEGIN's response-level binary labels as related to, but distinct from, the
sentence/claim-level attribution of the main constructs, and do not pool them.
\begin{table}[t]
\centering\small
\setlength{\tabcolsep}{3.2pt}
\begin{tabular}{lcccc}
\toprule
Metric & cmu-dog & TopicalChat & Wow & pooled \\
\midrule
lexical    & 0.743 & 0.668 & 0.846 & 0.766 \\
sem(MiniLM)& 0.747 & 0.565 & 0.826 & 0.756 \\
sem(MPNet) & 0.772 & 0.616 & 0.870 & 0.780 \\
combined   & 0.760 & 0.640 & 0.860 & 0.764 \\
BERTScore  & 0.762 & 0.833 & 0.845 & 0.799 \\
NLI$_{\mathrm{fe}}$ & 0.806 & 0.687 & 0.896 & 0.799 \\
NLI$_{\mathrm{cl}}$ & 0.880 & 0.756 & \textbf{0.944} & \textbf{0.890} \\
MiniCheck  & \textbf{0.922} & \textbf{0.877} & 0.885 & 0.883 \\
\bottomrule
\end{tabular}
\caption{\textbf{BEGIN dialogue-attribution} \citep{dziri2022begin} AUROC (knowledge-grounded
dialogue; $n{=}11{,}466$, $2{,}878$ attributable; best per column in bold). In this
\emph{new domain}, the grounding-trained family (NLI$_{\mathrm{cl}}$, MiniCheck) is
uniformly strongest, in contrast to the QA generated-answer attribution of
Table~\ref{tab:construct_audit}(a) where NLI$_{\mathrm{cl}}$'s rank was unstable. BEGIN is a
boundary condition --- a dataset where these metrics rank at the top across all three
sub-domains --- which is consistent
with (not a contradiction of) non-portability: the same NLI$_{\mathrm{cl}}$ collapses on LFQA.}
\label{tab:begin}
\end{table}

\section{Independent Replication and Robustness}
\label{app:replication}
\textbf{Independent multi-dataset replication.} To test whether non-transfer holds beyond
AttributionBench, we score the same zoo on eight \emph{disjoint} LLM-AggreFact source
datasets (excluding the two, ExpertQA and LFQA, that overlap our audit; $n{=}7336$ pairs).
No metric transfers there either (Table~\ref{tab:replication}): four different metrics win across
the eight datasets and the same clean MNLI scorer swings AUROC $0.494\!\to\!0.946$ --- the winning
metric changes by dataset, the signature of a genuine crossover rather than a uniform
difficulty shift.
\begin{table}[t]
\centering\footnotesize
\setlength{\tabcolsep}{3.2pt}
\resizebox{\columnwidth}{!}{%
\begin{tabular}{@{}lcccccc@{}}
\toprule
Dataset & Lex & Comb & BERTSc & NLI$_{\mathrm{cl}}$ & NLI$_{\mathrm{fe}}$ & MiniChk \\
\midrule
AggreFact-CNN    & 0.723 & 0.730 & \textbf{0.771} & 0.584 & 0.611 & 0.672 \\
AggreFact-XSum   & 0.500 & 0.548 & 0.543 & 0.625 & 0.714 & \textbf{0.784} \\
ClaimVerify      & 0.603 & 0.636 & 0.693 & 0.605 & 0.655 & \textbf{0.764} \\
FactCheck-GPT    & 0.605 & 0.742 & 0.727 & 0.819 & 0.828 & \textbf{0.833} \\
Reveal           & 0.597 & 0.813 & 0.661 & \textbf{0.946} & 0.923 & 0.944 \\
TofuEval-MediaS  & 0.605 & 0.641 & \textbf{0.695} & 0.675 & 0.635 & 0.686 \\
TofuEval-MeetB   & 0.659 & \textbf{0.685} & 0.642 & 0.552 & 0.610 & 0.674 \\
Wice             & 0.538 & 0.504 & 0.549 & 0.494 & 0.553 & \textbf{0.708} \\
\bottomrule
\end{tabular}}
\caption{\textbf{Independent replication: no audited metric transfers across a second,
disjoint multi-dataset benchmark.} AUROC against \emph{human} support labels on eight
LLM-AggreFact \citep{tang2024minicheck} source datasets ($n{=}7336$ pairs), excluding the
two (ExpertQA, LFQA) that overlap our AttributionBench audit so the panel is independent;
best per row in bold. As in Table~\ref{tab:construct_audit}, no single audited scorer is
best across datasets: four different metrics win the eight rows --- BERTScore (2),
MiniCheck (4), clean MNLI (Reveal), and the lexical/semantic blend (Comb, TofuEval-MeetB) ---
the same clean MNLI scorer swings from AUROC $0.494$ (Wice, $\le$chance) to $0.946$
(Reveal), so the winning metric changes by dataset,
and committing to the lowest-regret single metric (MiniCheck) still leaves mean
leave-one-dataset-out regret $0.015$ (max $0.099$). This is a faithfulness/grounding
benchmark (claim-vs-document support), adjacent to but distinct from sentence-level
citation attribution; we report it as an independent \emph{boundary replication} and do
\emph{not} pool it with the AttributionBench audit.}
\label{tab:replication}
\end{table}

\textbf{Portability-criterion sensitivity.} The non-transfer conclusion does not hinge on the
CI-relative screen: an $\epsilon$-regret sweep, Kendall-$\tau$ rank-inversion, a
cluster-respecting permutation, paired best-vs-rest tests, and threshold-free
AUPRC/balanced-accuracy all agree (Table~\ref{tab:sensitivity}).
\begin{table}[t]
\centering\small
\setlength{\tabcolsep}{4pt}
\begin{tabular}{@{}l l@{}}
\toprule
\multicolumn{2}{@{}l}{\emph{(a) Portability-criterion sensitivity (AttrBench 4 sources)}} \\
\midrule
$\epsilon$-regret tolerance & \# metrics ``transfer'' \\
$\epsilon \in \{0, .01, .02, .05\}$ & \textbf{0} \\
$\epsilon = 0.10$ & 1 (MiniCheck) \\
\midrule
\multicolumn{2}{@{}l}{\emph{(b) Cross-dataset rank stability (Kendall $\tau$)}} \\
\midrule
mean $\tau$ over dataset pairs & $-0.20$ \\
AttributedQA vs.\ LFQA & $-0.64$ ($p{=}.031$) \\
\midrule
\multicolumn{2}{@{}l}{\emph{(c) Interaction, main-effect-controlled}} \\
\midrule
rank concordance (Kendall $W$) & $0.07$, $p{=}.029$ \\
residual perm.\ (difficulty fixed) & $p{=}.053$ \\
\;\; \emph{mis-spec.\ shuffle, ref.\ only} & $p{<}.002$ \\
\midrule
\multicolumn{2}{@{}l}{\emph{(d) Paired best-vs-rest on LFQA (BERTScore best)}} \\
\midrule
BERTScore $-$ clean MNLI & $+.378$ $[.24,.50]$, $p{<}.001$ \\
BERTScore $-$ MiniCheck & $+.058$ $[{-}.02,.14]$, $p{=}.127$ \\
\bottomrule
\end{tabular}
\caption{\textbf{The non-transfer conclusion does not hinge on the CI-relative criterion.}
Four independent readings agree on AttributionBench's four sources: (a) under an
$\epsilon$-regret screen no audited metric is within $\epsilon$ of the per-dataset oracle on
\emph{every} dataset for any $\epsilon\le0.05$; (b) metric rankings are uncorrelated-to-negative
across datasets (the AttributedQA/LFQA ranking literally inverts); (c) the crossover survives
controlling for dataset difficulty --- within-dataset rank concordance is near zero (Kendall
$W{=}0.07$, $p{=}.029$) and a residual permutation that holds each dataset's difficulty fixed
puts $35\%$ of the cross-cell variance in the interaction residual (borderline, $p{=}.053$; the looser exchangeability shuffle that
conflates difficulty with interaction gives $p{<}.002$ and is shown for reference only); (d) the headline LFQA reversal is decisive (BERTScore beats clean MNLI by
$0.378$). \textbf{Honestly}, ``no single metric transfers'' is robust, but the best metric is
not always \emph{uniquely} best: on LFQA, BERTScore vs.\ MiniCheck is not significant
($p=0.127$). All quantities are recomputed and match the
Table~\ref{tab:construct_audit} AUROCs.}
\label{tab:sensitivity}
\end{table}

\textbf{A descriptive correlate of the collapse.} Evidence length / sentence count tracks
where clean MNLI fails (Table~\ref{tab:mechanism}), consistent with single-premise entailment
degrading under cross-sentence aggregation \citep{laban2022summac}; we report this as
descriptive, not causal.
\begin{table}[t]
\centering\footnotesize
\setlength{\tabcolsep}{3pt}
\resizebox{\columnwidth}{!}{%
\begin{tabular}{@{}lrrccc@{}}
\toprule
Dataset & \,words & sent. & NLI$_{\mathrm{cl}}$ & BERTSc & MiniChk \\
\midrule
AttributedQA       & 103 & 4.2  & \textbf{0.904} & 0.692 & 0.884 \\
Stanford-GenSearch & 82  & 4.1  & 0.822 & 0.838 & 0.783 \\
ExpertQA           & 174 & 7.9  & 0.565 & 0.569 & 0.577 \\
LFQA               & 321 & 13.9 & 0.531 & \textbf{0.909} & 0.852 \\
\bottomrule
\end{tabular}}
\caption{\textbf{A descriptive correlate of the clean-MNLI collapse: evidence length /
sentence count.} Mean evidence length (words) and evidence sentence count per
AttributionBench source, against per-dataset AUROC. The clean MNLI scorer is strongest where
evidence is short and few-sentence (AttributedQA: 103 words, 4.2 sentences $\to$ $0.904$) and
collapses where evidence is long and many-sentence (LFQA: 321 words, 13.9 sentences $\to$
$0.531$), while token-level BERTScore moves the opposite way ($0.692\to0.909$). This is
consistent with a single-premise entailment head degrading when support must be aggregated
across many sentences \citep{laban2022summac}. We report this as a \textbf{descriptive}
correlate, not a causal mechanism: with four datasets the dataset-level association is
underpowered, the per-example abstractiveness signal does not cleanly separate NLI from
BERTScore, and ExpertQA is a near-chance exception. Identifying the causal driver is left to
future work.}
\label{tab:mechanism}
\end{table}

\textbf{Is a prompt-based LLM judge more stable across datasets?} We also run a prompt-based LLM judge
(Opus~4.8) as a scoring metric across the four AttributionBench sources (Table~\ref{tab:llm_judge}). On those
same four sources it never drops to the chance-level collapses the strong automatic scorers suffer (range
$0.731$--$0.918$ vs.\ clean MNLI's $0.531$--$0.904$ on the four) and does not collapse on LFQA, but is not uniformly best, costs
${\sim}100\times$ more, and is non-deterministic. Its complete-case AUROC uses $408$ of $480$ sampled items
with valid parsed scores. The other $72$ ($15\%$) are unscored; the saved scores do not distinguish
refusals, truncation, parsing failures, and API errors. Non-random exclusions could inflate
performance and its apparent stability advantage. Table~\ref{tab:llm_judge} reports counts and
a sensitivity check assigning $0.50$ to every unscored item: all four AUROC point estimates remain
above chance. This check does not resolve selection bias or establish superiority over automatic
scorers, which were evaluated on the full source datasets rather than the judge's retained subsets.
The judge remains subject to target-dataset validation.
\begin{table}[t]
\centering\footnotesize
\setlength{\tabcolsep}{3pt}
\begin{tabular}{@{}lrrrr@{}}
\toprule
Dataset & Valid $n$ & Judge & Imputed & Best auto. \\
\midrule
AttributedQA       & 101/120 & 0.918 & 0.892 & 0.904 \\
LFQA               & 105/120 & 0.878 & 0.859 & 0.909 \\
Stanford-GenSearch & 100/120 & 0.794 & 0.763 & 0.838 \\
ExpertQA           & 102/120 & 0.731 & 0.708 & 0.618 \\
\bottomrule
\end{tabular}
\caption{\textbf{LLM-judge AUROC and unscored-item sensitivity.} Claude Opus~4.8 emits a graded support
score evaluated against \emph{human} AttributionBench labels. Sampling used $60$ positive and $60$
negative items per source; $408/480$ yielded valid scores. Judge: complete-case AUROC. Imputed:
AUROC on all $120$/source after assigning $0.50$ to unscored items. Exclusions are $19/15/20/18$
in row order. Above-chance point estimates persist under this imputation, which does not rule out
selection bias. Best auto.: best of the eight automatic scorers on each \emph{full} source dataset
(Table~\ref{tab:construct_audit}), not a paired comparison on the judge subset: clean MNLI, BERTScore,
BERTScore, and MPNet in row order. The judge is not uniformly best by these point estimates and remains
costlier (${\sim}100\times$) and non-deterministic. See App.~\ref{app:replication} for limits.}
\label{tab:llm_judge}
\end{table}

\textbf{LLM-judge prompt templates.} For reproducibility we give the exact prompts verbatim
(Figure~\ref{fig:prompts}). The judge runs as a \emph{scoring metric}:
\texttt{claude-opus-4-8}, API-default decoding (no temperature parameter supplied),
\texttt{maxTokens}$=20$, evidence/claim/question truncated to $1500$/$600$/$300$ characters; the
support probability is parsed with a regex and clamped to $[0,1]$. Unscored items are
excluded from the primary AUROC and imputed only for the sensitivity check. The separate cross-family \emph{annotation} probe
(Section~\ref{sec:benchmark2}; Opus~4.8 and GPT-5.4, $n{=}160$) instead elicits a discrete label,
scored as inter-annotator agreement, never as gold.

\begin{figure*}[t]
\centering
\begin{minipage}{0.49\textwidth}
\footnotesize
\textbf{(a) Graded scoring prompt} (LLM judge as a metric):
\begin{verbatim}
You are an automated ATTRIBUTION-scoring
rubric for a QA system. Given a CLAIM (a
sentence from a generated answer) and the
EVIDENCE it cites, rate how fully the claim
is supported by (attributable to) the
evidence.

Output EXACTLY one line and nothing else,
even if uncertain:
SUPPORT: <number 0.00-1.00>

1.00 = every factual element of the claim is
directly supported by the evidence; 0.00 =
the evidence is unrelated to or contradicts
the claim; intermediate values for partial
support. Use the full range and be
calibrated. This is an automated rubric:
always return a number (use 0.50 if truly
unsure). Do not refuse and do not explain.

QUESTION: {question}
CLAIM:    {claim}
EVIDENCE: {evidence}
\end{verbatim}
\end{minipage}\hfill
\begin{minipage}{0.49\textwidth}
\footnotesize
\textbf{(b) Discrete annotation prompt} (cross-family $\kappa$ probe):
\begin{verbatim}
You are judging ATTRIBUTION for a question-
answering system. Given a CLAIM (a sentence
from a generated answer) and the EVIDENCE it
cites, decide: is the claim fully supported
by (attributable to) the evidence?

Answer STRICTLY in this format on one line:
LABEL: <attributable|not_attributable>
CONF:  <0-1>

A claim is "attributable" only if all of its
factual content is supported by the evidence.
If the evidence is unrelated, contradicts, or
only partially supports the claim, answer
not_attributable.

QUESTION: {question}
CLAIM:    {claim}
EVIDENCE: {evidence}
\end{verbatim}
\end{minipage}
\caption{\textbf{Verbatim LLM prompt templates.} (a) The graded support-probability prompt used to run
Claude Opus~4.8 as a scoring metric (Table~\ref{tab:llm_judge}); (b) the discrete-label prompt used by the
$n{=}160$ cross-family adjudication probe (Opus~4.8 and GPT-5.4). \texttt{\{question\}}, \texttt{\{claim\}},
and \texttt{\{evidence\}} are filled per item. Both are disclosed as LLM outputs, never human/gold.}
\label{fig:prompts}
\end{figure*}

\textbf{Cross-benchmark landscape.} Pooling per-dataset AUROCs across all \textbf{17} human-labeled datasets
we audit (Table~\ref{tab:meta}), six different metrics are best on at least one dataset, the most frequent
winner (MiniCheck) is best on only $6/17$ ($35\%$), and every metric's AUROC swings by $0.34$--$0.45$ across
datasets. (Descriptive landscape; constructs are tagged and not pooled into a single test.)
\begin{table}[t]
\centering\small
\setlength{\tabcolsep}{4pt}
\begin{tabular}{lccc}
\toprule
Metric & min & max & range \\
\midrule
lexical Jaccard      & 0.500 & 0.839 & 0.339 \\
MiniLM cosine        & 0.499 & 0.841 & 0.342 \\
MPNet cosine         & 0.477 & 0.851 & 0.374 \\
combined blend       & 0.504 & 0.845 & 0.341 \\
BERTScore            & 0.543 & 0.909 & 0.366 \\
NLI$_{\mathrm{fe}}$ (FEVER) & 0.553 & 0.923 & 0.370 \\
NLI$_{\mathrm{cl}}$ (clean MNLI) & 0.494 & 0.946 & \textbf{0.452} \\
MiniCheck            & 0.577 & 0.944 & 0.367 \\
\bottomrule
\end{tabular}
\caption{\textbf{Cross-benchmark landscape: every audited metric swings widely across datasets,
and no metric dominates.} Per-metric AUROC range across the \textbf{17 human-labeled datasets} we
audit, spanning four evaluation settings (5 generated-answer attribution incl.\ HAGRID; 1
fact-check; 3 dialogue-attribution sub-domains; 8 LLM-AggreFact faithfulness datasets). Six
different metrics are best on at least one dataset; the most frequent winner (MiniCheck) is best on
only $6/17$ ($35\%$). The same clean MNLI scorer spans AUROC $0.494$ (below chance) to $0.946$ --- a
$0.45$ swing. The best fixed default (MiniCheck, mean AUROC $0.791$) still leaves mean
leave-one-dataset-out regret $0.026$ (max $0.099$). This is a \emph{descriptive} landscape pooling
the per-dataset AUROCs reported in the paper; datasets span distinct constructs and are
\emph{not} pooled into a single significance test.}
\label{tab:meta}
\end{table}

\end{document}